\documentclass{sig-alternate-2013}

\usepackage{epstopdf}
\usepackage{url}
\usepackage{textcomp}
\usepackage[english]{babel}
\usepackage[autostyle]{csquotes}

\newfont{\mycrnotice}{ptmr8t at 7pt}
\newfont{\myconfname}{ptmri8t at 7pt}
\let\crnotice\mycrnotice%
\let\confname\myconfname%

\permission{Permission to make digital or hard copies of all or part of this work for personal or classroom use is granted without fee provided that copies are not made or distributed for profit or commercial advantage and that copies bear this notice and the full citation on the first page. Copyrights for components of this work owned by others than ACM must be honored. Abstracting with credit is permitted. To copy otherwise, or republish, to post on servers or to redistribute to lists, requires prior specific permission and/or a fee. Request permissions from Permissions@acm.org.}
\conferenceinfo{MM'15,}{October 26--30, 2015, Brisbane, Australia.}
\copyrightetc{\copyright~2015 ACM. ISBN \the\acmcopyr}
\crdata{978-1-4503-3459-4/15/10\ ...\$15.00.\\
DOI: http://dx.doi.org/10.1145/2733373.2806293}

\begin{document}
%

\title{Deep Multimodal Speaker Naming
}
%
%
%
%
%

\numberofauthors{6} 
%
\author{
%
%
\alignauthor
Yongtao Hu\\
       \affaddr{The University of Hong Kong}\\
       \email{ythu@cs.hku.hk}
\alignauthor
Jimmy SJ. Ren\\
       \affaddr{SenseTime Group Limited}\\
       \email{rensijie@sensetime.com}
\alignauthor
Jingwen Dai\\
       \affaddr{Xim Industry Inc.}\\
       \email{dai@ximmerse.com}
\and  
\alignauthor
Chang Yuan\\
       \affaddr{Lenovo Group Limited}\\
       \email{yuanchang@lenovo.com}
\alignauthor
Li Xu\\
       \affaddr{SenseTime Group Limited}\\
       \email{xuli@sensetime.com}
\alignauthor
Wenping Wang\\
       \affaddr{The University of Hong Kong}\\
       \email{wenping@cs.hku.hk}
}


\maketitle
\begin{abstract}
Automatic speaker naming is the problem of localizing as well as identifying each speaking character in a TV/movie/live show video. This is a challenging problem mainly attributes to its multimodal nature, namely face cue alone is insufficient to achieve good performance. Previous multimodal approaches to this problem usually process the data of different modalities individually and merge them using handcrafted heuristics. Such approaches work well for simple scenes, but fail to achieve high performance for speakers with large appearance variations. In this paper, we propose a novel convolutional neural networks (CNN) based learning framework to automatically learn the fusion function of both face and audio cues. We show that without using face tracking, facial landmark localization or subtitle/transcript, our system with robust multimodal feature extraction is able to achieve state-of-the-art speaker naming performance evaluated on two diverse TV series. The dataset and implementation of our algorithm are publicly available online.
\end{abstract}

\category{H.3}{Information Storage and Retrieval}{Content Analysis and Indexing}


\keywords{Speaker Naming, Multimodal, CNN, Deep Learning}

\section{Introduction}
Identifying speakers, or speaker naming (SN), in movies, TV series and live shows is
a fundamental problem in many high-level video analysis tasks, such as semantic indexing and retrieval \cite{zhang2013attribute}
and video summarization \cite{takenaka2012drive}, etc. As noted by previous authors \cite{Everingham2006}, automatic SN is extremely
challenging as characters exhibit significant variation of visual
appearance due to changes in scale, pose, illumination, expression, dress, hair style, etc. Additional problems with video acquisition, such as poor image quality and motion blur, make the matter even worse. Previous studies using only a single visual cue, such as face features, failed to generate satisfactory results.

Real-life TV/movie/live show videos are all multimedia data consisting of multiple sources
of information. In particular, audio provides reliable supplementary information for SN task because it is closely associated with the video. In this paper, we propose a novel CNN based learning framework to tackle the SN problem. Unlike previous methods which investigated different modalities individually, our method automatically learns the fusion function of both face and audio cues and outperforms other state-of-the-art methods without using face/person tracking, facial landmark localization or subtitle/transcript. Our system is also trained end to end, providing an effective way to generate high quality intermediate unified features to distinguish outliers.

\begin{figure*}[t]
\begin{center}
\includegraphics[width=0.93\linewidth]{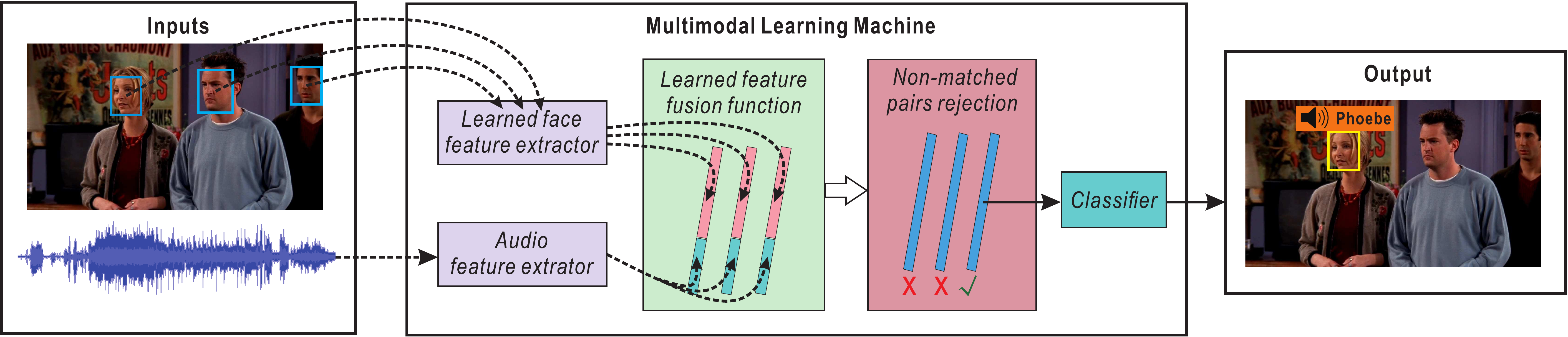}
\end{center}
\vspace{-1.2em}
   \caption{Multimodal learning framework for speaker naming.}
\label{fig:framework}
\end{figure*}

\vspace{3pt}
{\textbf{Contributions.}}\quad
1) a novel CNN based framework which automatically learns high quality multimodal feature fusion functions; 2) a systematic approach to reject outliers for multimodal classification tasks typified by SN, and 3) a state-of-the-art system for practical SN applications.



\section{Related Work}
\label{sec::RelatedWork}

Automatic SN in TV series, movies and live shows has received increasing attention in the past decade.
In previous works like \cite{liu2008naming}, SN was considered
as an automatic face recognition problem.
Recently, more researchers have tried to make use of video context to boost performance. Most of these works focused on \emph{naming face tracks}.
In \cite{Everingham2006}, cast members are automatically labelled
by detecting speakers and aligning subtitles/transcripts to obtain identities.
This approach had been adapted and further refined by ~\cite{Sivic2009}.
Bauml \textit{et al.} \cite{Bauml2013} use a similar
method to automatically obtain labels for those face tracks that can be detected as speaking.
However, these labels are typically noisy and incomplete (i.e., usually only $20$-$30\%$ of the tracks can be assigned a name) \cite{Bauml2013}.
That is mainly due to that speaker detection relies heavily on lip movement detection,
which is not reliable for videos of low quality or with large face pose variation.

In \cite{Tapaswi2012}, each TV series episode is modeled as a Markov Random Field, which integrates face
recognition, clothing appearance, speaker recognition and contextual constraints in a probabilistic manner.
The identification task is then formulated as an energy minimization problem.
In \cite{yang2004naming, yang2005multiple}, person naming is resolved by a statistical learning or multiple instances learning framework.
Bojanowski \textit{et al.}~\cite{Bojanowski2013} utilize scripts as weak supervision to learn a joint model of
actors and actions in movies for character naming.
Although these methods try to solve character naming or SN problem in new machine learning frameworks,
they still heavily rely on accurate face/person tracking, motion detection, landmark detection and aligned transcripts or captions.

Unlike all these previous works, our approach does not rely on face/person tracking, motion detection, facial landmark localization or subtitle/aligned transcript as well as handcrafted features engineering. With only the input of cropped face regions and corresponding audio segment, our approach
recognizes speaker in each frame in real-time.

\section{Multimodal CNN Framework}
\label{sec::Method}

Our approach is a learning based system in which we fuse the face and audio cues in the feature extraction level. The face feature extractor is learned from data rather than handcrafted. Then our learning framework is able to leverage both face and audio features and learns a unified multimodal feature extractor. This enables a larger learning machine to learn a unified multimodal classifier which takes both face image and speaker's sound track as inputs.
The overview of the learning framework is illustrated in Figure \ref{fig:framework}.

\subsection{Multimodal CNN Architecture}
We adopted CNN \cite{KrizhevskySH12} as the baseline model in our learning machine. As we will see shortly, CNN's architecture is inherently extensible. This makes our extension to multimodal learning concise, efficient but powerful.

The role of CNN in our framework is two-fold. Firstly, it learns a face feature extractor from face imagery data so that we have a solid face recognition baseline. Secondly, it combines both face feature extractor as well as the audio feature extractor and learns a unified multimodal classifier.

\begin{figure}[!htb]
  \centering
  \includegraphics[width=0.95\linewidth]{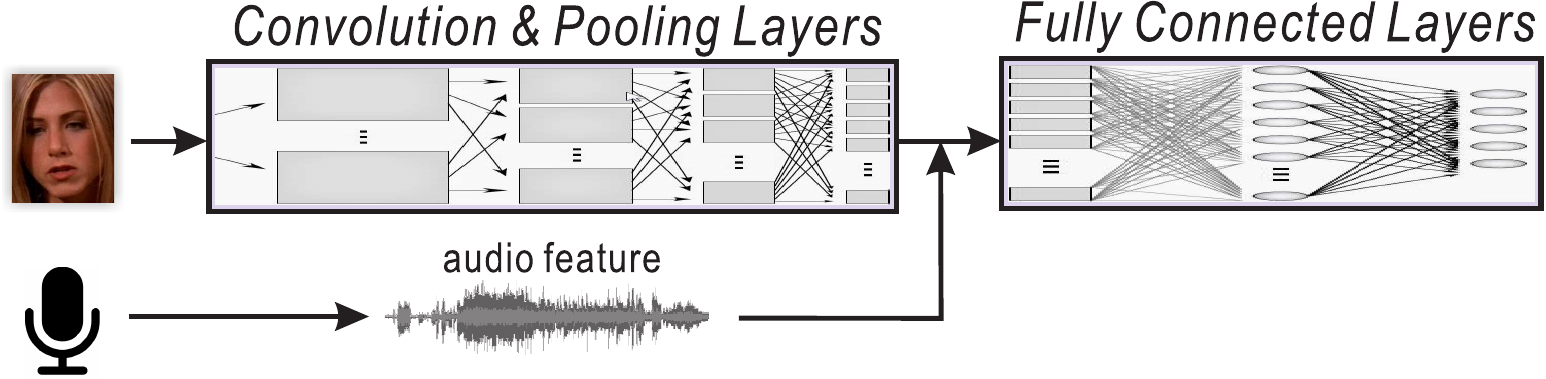}
  \vspace{-1.2em}
  \caption{Multimodal CNN architecture.}
  \label{fig:fig2} 
\end{figure}

Figure~\ref{fig:fig2} illustrates the design of our model. We will later show with insights that this model is very effective for SN tasks despite its conciseness.

In the trainable face feature extractor part, each layer of the network can be expressed as

\begin{equation}
N_c(\mathcal{I})=\sigma(\mathcal{P}(\sigma(\mathcal{I} * K^l + b^l))), \; l = 1, 2, ... , n,
\label{Eq:CNN1}
\end{equation}

\noindent where $\mathcal{I}$ is the input for each layer. $\mathcal{I}$ is usually a 3D image volume, namely 3-channel input images when $l=1$, multi-channel feature maps when $1 < l \leq n$. $K^l$ and $b^l$ are the trainable convolution kernels and trainable bias term in layer $l$ respectively. $\sigma$ represents the nonlinearity in the network, which is modeled by a rectifier expressed as $f(x) = \max(0, x)$. $\mathcal{P}$ is a pooling function which subsamples the inputs by a factor of 2. Same nonlinearity is applied after the pooling function. When $l=n$ the output of $N_c(\mathcal{I})$ is a one dimensional high level feature vector.

For audio feature extraction, we use mel frequency cepstral coefficients (MFCCs) \cite{sahidullah2012design}. The MFCCs of one audio frame is also an one dimensional feature vector. This allows us to ensemble a unified multimodal feature by stacking $N_c(\mathcal{I})$ and MFCCs together.

It is worth noting that stacking of face feature and MFCCs in this stage is non-trivial in terms of classification. The reason is the ensuing trainable classifier essentially learns a higher dimensional nonlinear feature representation of the previous layer by mapping the stacked multimodal feature to a higher dimensional feature space. This is expressed as
\begin{equation}
N_f(\mathcal{F})=\sigma(\mathcal{F} \cdot \mathcal{W}^l + b^l), \; l = n+1, n+2, ... , m,
\label{Eq:CNN2}
\end{equation}

\noindent where $\mathcal{F}$ is the stack of face feature and MFCCs with layer $l=n+1$. When $n+1 < l < m$, we impose constraint $Dim(\mathcal{F}^{l-1}) < Dim(\mathcal{F}^{l})$ where $Dim()$ denotes the dimension of the intermediate feature vector, which promotes the learning of higher dimensional feature mapping. Such feature mapping is realized by the trainable weights $\mathcal{W}$ and $b$ as well as the nonlineary $\sigma$. The system outputs the decision values of each class label by going through a softmax layer when $l=m$. The cross-entropy error function $\sum_{i=0}^{n} \ln(o_i) \cdot t_i$ is used as the error function during training, where $o_i$ is the $i$-th element in $N_f(\mathcal{F})$, $t$ is the ground truth class label. Though the conciseness of the model, one key insight of this approach is the whole system is trained end to end such that the influence of face feature extractor and MFCCs to the whole network is interwinding through learning.

\vspace{4pt}
{\textbf{Multimodal Feature Extraction.}}\quad
One important character of the CNN based classifier is its intermediate layers are essentially high level feature extractors. Previous studies \cite{Oquab14} showed that such high level features is very expressive and can be applied in tasks such as recognition and content retrieval. It was not clear if such high level feature extraction mechanism works well in the context of multimodal learning. We will show in our experiments that our method is able to generate high quality multimodal features which is highly expressive in distinguishing outlier samples. This discovery forms one of the most important building blocks of making our system superior for real-life SN applications.

\section{Experiments}
\label{sec::Exp}

\textbf{Experimental Setup.}\quad
We evaluate our framework on over three hours videos of nine episodes from two TV series, i.e. \enquote{Friends} and \enquote{The Big Bang Theory} (\enquote{BBT}). For \enquote{Friends}, faces and audio from \textit{S01E03} (Season 01, Episode 03), \textit{S04E04}, \textit{S07E07} and \textit{S10E15} serve as the training set and those from \textit{S05E05} as the evaluation set. Note that, the whole \enquote{Friends} TV series of ten seasons is taken over a large time range of ten years. To leverage such a long time span, we intentionally selected these five episodes that spans the whole range. For \enquote{BBT}, as in \cite{Tapaswi2012}, \textit{S01E04}, \textit{S01E05} and \textit{S01E06} are for training and \textit{S01E03} for testing. For these two TV series, we only report performance of the leading roles, including six ones of \enquote{Friends}, i.e. \textit{Rachel}, \textit{Monica}, \textit{Phoebe}, \textit{Joey}, \textit{Chandler} and \textit{Ross}, and five ones of \enquote{BBT}, i.e. \textit{Sheldon}, \textit{Leonard}, \textit{Howard}, \textit{Raj} and \textit{Penny}.

We conduct three experiments in terms of 1) face recognition; 2) identifying non-matched face-audio pairs and 3) real world SN respectively. For face recognition using both face and audio information, we only identify matched face-audio pairs. We further show how our model be able to classify matched face-audio pairs from non-matched ones. It is worth noting that the first two experiments provide solid foundations towards achieving promising performance in our third real world SN experiment. It also justifies the effectiveness of the building blocks in our resulting system.

Our CNN's detailed setting is described as follows. The network has $2$ alternating convolutional and pooling layers in which the sizes of the convolution filters used are $15 \times 15$ and $5 \times 4$ respectively. The connection between the last pooling layer and the fully connected layer uses filters of size $7 \times 5$. The number of feature maps generated by the convolutional layers are $48$ and $256$ respectively. For fully connected layers, the number of hidden units are $1,024$ and $2,048$ respectively. Such architecture requires more than $11$ million trainable parameters. All the bias terms are initialized to $0.01$ to prevent the dead unit caused by rectifier units during training. All other parameters are firstly initialized within the range of $-1$ to $1$ drawn from a Gaussian distribution and then scaled by the number of fan-ins of hidden unit they connect to. Average pooling of factor $2$ is used throughout the network.

\subsection{Face Model}
We evaluate our model for face recognition on \enquote{Friends} (all face images resized to $50\times40$). We also test four previous algorithms under the same setting, i.e. Eigenface \cite{turk1991eigenfaces}, Fisherface \cite{belhumeur1997eigenfaces}, LBP \cite{ahonen2006face} and OpenBR/4SF \cite{klontz2013open}.

Accuracies of these four previous methods are $60.7\%$, $64.5\%$, $65.6\%$ and $66.1\%$ respectively. All four previous algorithms fail to work well (all $<70\%$), on the other hand, our method works better for every subject and achieves an accuracy of $86.7\%$. The results are expected as previous algorithms either require alignment of the face images or detecting facial feature points or both. This makes them not able to work well in the small sized face images that are extracted from unconstrained videos, which has no guarantee of alignment of the images, challenging large variations in pose, illumination and aging, etc.

We further apply audio to fine-tune our face model. The weights in this extended network is initialized by the parameters in the face-alone network. For the newly introduced parameters by the audio inputs, they are initialized in the same way as presented before. Concerning audio features, a window size of 20ms and a frame shift of 10ms are used. We then select mean and standard deviation of 25D MFCCs, and standard deviation of 2-$\Delta$MFCCs, resulting in a total of 75 features per audio sample.
For each face, we catenate it with 5 audio samples of the same subject that are randomly selected to generate face-audio pairs.

\begin{figure}[t]
\begin{center}
\includegraphics[width=0.9\linewidth]{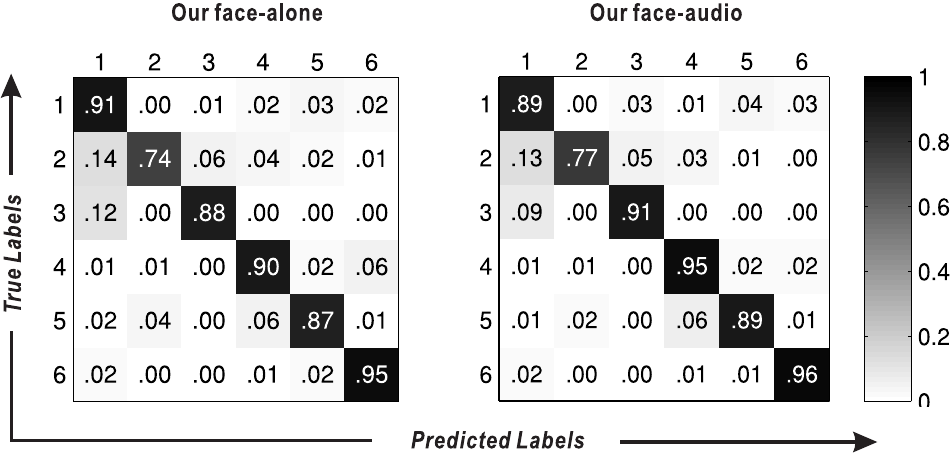}
\end{center}
\vspace{-1.2em}
   \caption{Confusion matrices of our face-alone and face-audio models for face recognition on \enquote{Friends}. Labels 1-6 stand for the six subjects accordingly, i.e. \textit{Rachel}, \textit{Monica}, \textit{Phoebe}, \textit{Joey}, \textit{Chandler} and \textit{Ross}.}
\label{fig:face-result}
\vspace{-0.5em}
\end{figure}

Compared with previous face-alone model (acc: $86.7\%$), our face-audio model further improved this to $88.5\%$ with corresponding confusion matrix shown in Figure \ref{fig:face-result}. We can clearly see that, by adding audio information to the model, the accuracies of identifying all the subjects improve by $1$-$5\%$ except a slight drop for \textit{Rachel}.

\subsection{Identifying Non-matched Pairs}
In above experiments, all face-audio samples are matched pairs, i.e. belong to the same person. However, this condition cannot be fulfilled in practice. Consider a speaking frame, there are $N$ faces, one of which is speaking, see Figure \ref{fig:framework} as an example where $N=3$. In order to find the correct speaker, we need to examine all face-audio pairs. All the pairs are non-matched except the one of the real speaker. And, it is almost impossible to train all possible non-matched pairs because new faces are unpredictable.

Thus, to identify non-matched pairs, a better way is to develop new strategies at the same time guarantee the quality of the face model. Instead of using the final output label of our face models, we explore the effectiveness of the features returned from the model in the last layer. As baseline, we train two binary supporting vector machines (SVM) \cite{CC01a}. One is trained on the $1024$D fused feature that returned from our face-audio model and the other trained on $1024$D face feature returned from our face-alone model concatenating with $75$D audio feature (MFCC). We then train another SVM model using the same setting with the second SVM expect that we replace the $1024$D face feature by the same dimensional fused feature from our face-audio model.

We test these three models on the evaluation video, which contains in total $17,131$ speaking frames. It will count as correct if the most confident
face-audio pair matches, i.e. both from the same person. Two baseline SVMs achieve $82.2\%$ and $82.9\%$ respectively, whilst the third one can achieve $84.1\%$. Results clearly justify that the fused feature is more discriminative than the original face feature. On the other hand, we believe it also shows that the fused feature and MFCCs capture different but complimentary dimensions of the required information in distinguishing non-matched pairs.

\subsection{Speaker Naming}
The goal of speaker naming is to identify the speaker in each frame, i.e. find out the matched face-audio pair and also identify it. It's worth noting that such a problem can be viewed as an extension of previous experiment of identifying non-matched pairs (reject all non-matched ones).

\begin{figure}[t]
\begin{center}
\vspace{-1.0em}
\includegraphics[width=0.9\linewidth]{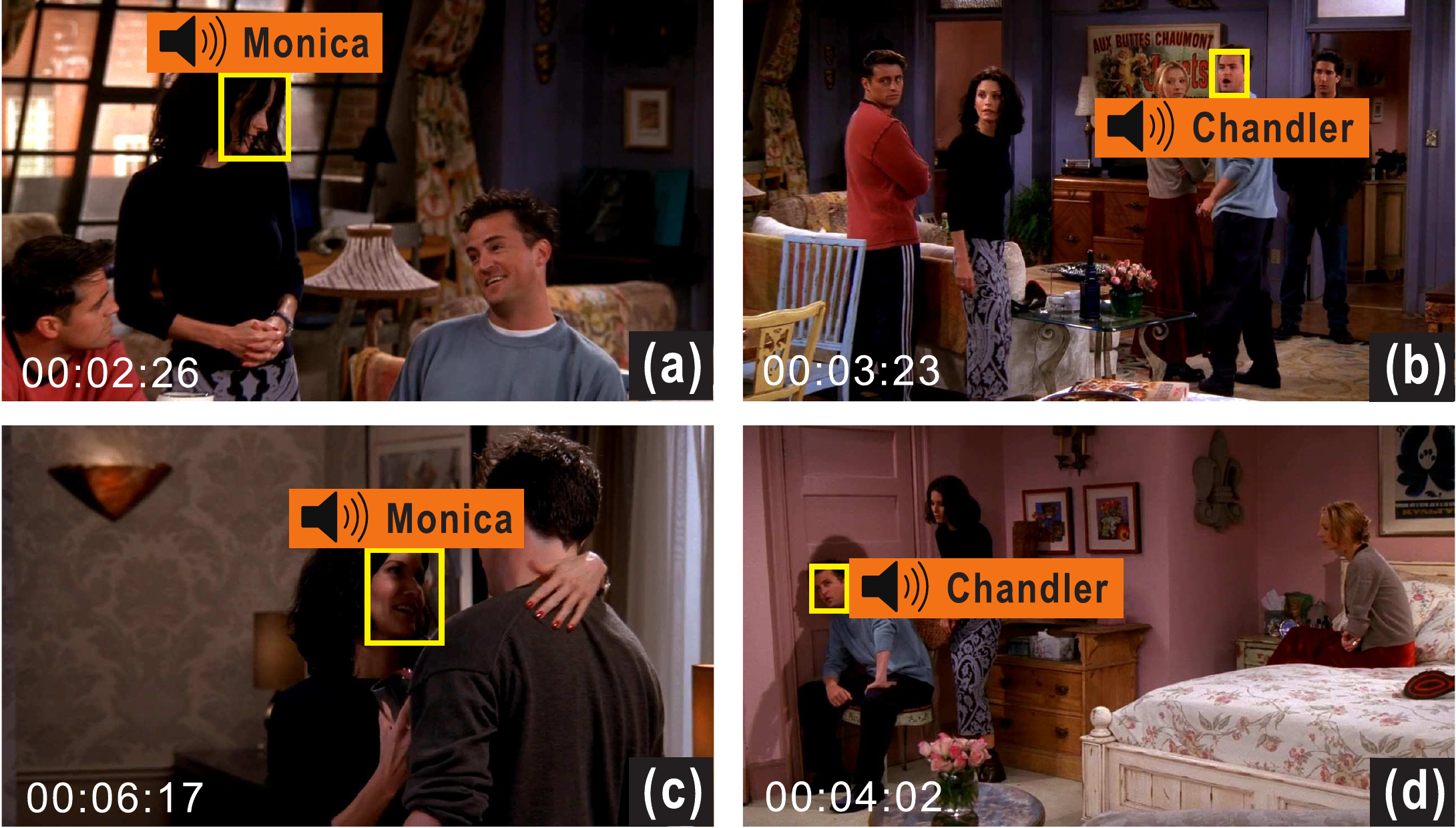}
\end{center}
\vspace{-1.6em}
   \caption{Speaker naming result under various conditions, including pose (a)(c)(d), illumination (c)(d), small scale (b)(d), occlusion (a) and clustered scene (b)(d), etc (time stamp shown at the bottom left).}
\label{fig:result}
\vspace{-0.7em}
\end{figure}

For the in total $17,131$ speaking frames in the evaluation video \textit{Friends.S05E05}, we applied the third SVM to reject all non-matched pairs. The remaining pair will be assigned with the label returned by our face-audio model. Under such setting, we can achieve the SN accuracy of $90.5\%$. Sample SN result can be viewed from Figure \ref{fig:result}.

{\textbf{Compared with Previous Works.}}\quad
Previous works \cite{Bauml2013,Tapaswi2012} have addressed similar SN problem by incorporating face, facial landmarks, cloth features, character tracking and associated video subtitles/transcripts. They evaluated on \enquote{BBT} and achieved SN accuracy of $77.8\%$ and $80.8\%$ respectively (evaluation on \textit{S01E03}). In comparison,
we can achieve SN accuracy of $82.9\%$ without introducing any face/person tracking, facial landmark localization or subtitle/transcript.

\subsection{Applications}

Speaking activity is the key of multimedia data content. With our system, detailed speaking activity can be obtained, including speakers' locations, identities and speaking time ranges, etc, which further enables many useful applications. We highlight two major applications in the following (please refer to our supplementary video for details):

{\textbf{Video Accessibility Enhancement.}}\quad
With speakers' locations, we can generate on-screen dynamic subtitles next to the respective speakers thus enhance video accessibility for the hearing impaired \cite{hong2010dynamic} and enhance the overall viewing experience as well as reduce eyestrain for normal viewers \cite{hu2014speaker}.

\begin{figure}[!ht]
\begin{center}
\vspace{-1.0em}
\includegraphics[width=\linewidth]{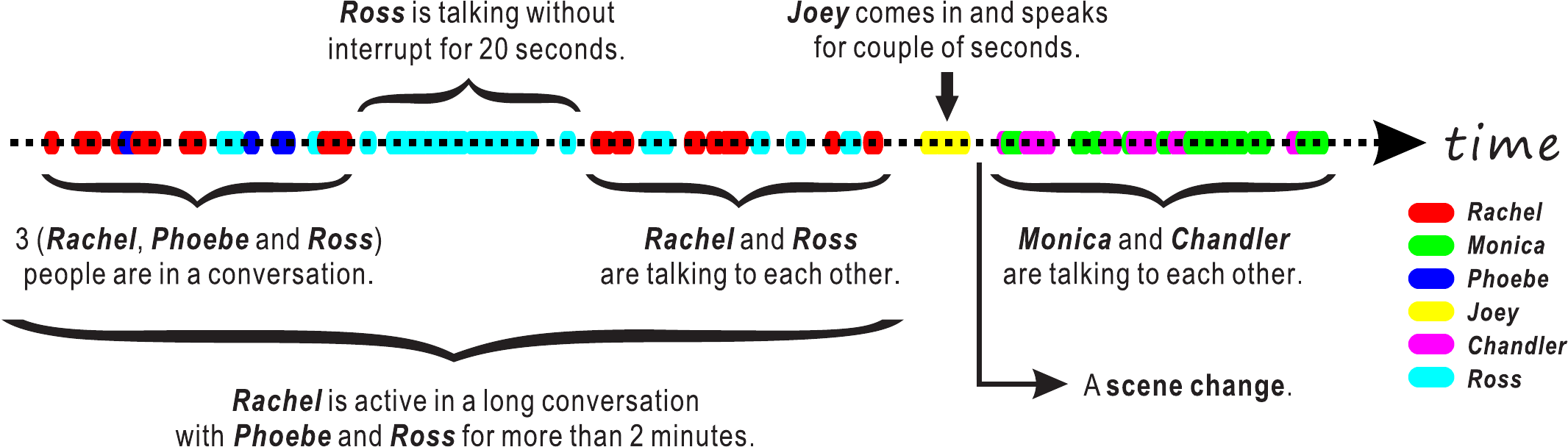}
\end{center}
\vspace{-1.6em}
   \caption{Speaking activity and video summarization for a 3.5 minutes video clip of \textit{Friends.S05E05}.}
\label{fig:video-summary}
\vspace{-0.7em}
\end{figure}

{\textbf{Multimedia Data Retrieval and Summarization.}}\quad
With the detailed speaking activity, we can further achieve some high-level multimedia data summarization tasks, including characters conversation information and scene changing information, etc, based on which fast video retrieval is possible. We highlight such information in Figure \ref{fig:video-summary}.

\section{Conclusions}
\label{sec::Concl}
In this paper, we propose a CNN based multimodal learning framework to tackle the task of \textit{speaker naming}. Our approach is able to automatically learn the fusion function of both face and audio cues. We show that our multimodal learning framework not only obtains high face recognition accuracy but also extracts representative multimodal features which is the key to distinguish sample outliers. By combining the aforementioned capabilities, our system achieved state-of-the-art performance on two diverse TV series without introducing any face/person tracking, facial landmark localization or subtitle/transcript. The dataset and implementation of our algorithm, based on VCNN \cite{ren2015vectorization}, are publicly available online at \url{http://herohuyongtao.github.io/research/publications/speaker-naming/}.



%
{\fontsize{8}{8.2}\selectfont
\bibliographystyle{abbrv}
\bibliography{sigproc}  
}
\end{document}